%% file: root.tex
\documentclass[letterpaper, 10 pt, conference]{ieeeconf}  

\IEEEoverridecommandlockouts                              

\usepackage{graphicx}
\usepackage{caption}
\usepackage{subcaption}

\title{\LARGE \bf
Self-supervised Pre-training for Semantic Segmentation in an Indoor Scene
}

\author{Sulabh Shrestha$^{1}$, Yimeng Li$^{1}$ and Jana Ko\v{s}ecka$^{1}$
\thanks{$^{1}$Sulabh Shrestha, Yimeng Li and Jana Ko\v{s}ecka are with the 
        Department of Computer Science, 
        George Mason University, 
        4400 University Dr, Fairfax, VA, USA
        {\tt\small \{sshres2,yli44,kosecka\}@gmu.edu}
        }%
}

\begin{document}

\maketitle
\thispagestyle{empty}
\pagestyle{empty}


\begin{abstract}
    The ability to endow maps of indoor scenes with semantic information is an integral part of robotic agents which perform different tasks such as target driven navigation, object search or object rearrangement. The state-of-the-art methods use Deep Convolutional Neural Networks (DCNNs) for predicting semantic segmentation of an image as useful representation for these tasks. 
    The accuracy of semantic segmentation depends on the availability and the amount of labeled data from the target environment or the ability to bridge the domain gap between test and training environment.
    We propose RegConsist, a method for self-supervised pre-training of a semantic segmentation model, exploiting the ability of the agent to move and register multiple views in the novel environment. Given the spatial and temporal consistency cues used for pixel level data association, we use a variant of contrastive learning to train a DCNN model for predicting semantic segmentation from RGB views in the target environment. The proposed method outperforms models pre-trained on ImageNet and achieves competitive performance when using models that are trained for exactly the same task but on a different dataset. We also perform various ablation studies to analyze and demonstrate the efficacy of our proposed method.

\end{abstract}


\input{tex_files/Introduction}


\input{tex_files/RelatedWork}


\input{tex_files/Method}


\section{Experiments} \label{sec:experiments}

\subsection{Dataset} \label{sec:exp-data}


We perform our experiments on Active Vision Dataset (AVD) \cite{Ammirato_ICRA_2017_AVD} and Replica dataset \cite{replica19arxiv}. AVD is a real-world dataset that consists of scenes from different apartments and offices. Each scene contains images taken by a robot in a grid-like manner and a few of the images are annotated. We use \textit{Home\_006\_1} which contains 2412 images among which 43 images are annotated. Replica is a photo-realistic dataset that consists of multiple indoor environments. Both the datasets contain ground-truth depth as well as intrinsic and extrinsic parameters of the camera. Since AVD is a real-world dataset, it contains noisy and incomplete depth measurements. We start with Replica dataset to disambiguate the cause of errors between the noise in the data and our approach.  Unless otherwise stated, we experiment on \textit{frl\_apartment\_1} environment which has an apartment-like setting as shown in Figure \ref{fig:segmentation}. We use Habitat simulator \cite{habitat} to move the agent in the environment and generate overlapping views similar to AVD. 

We compare our model trained in self-supervised way with a segmentation model already trained on ADE20K dataset \cite{ade20k,ade20kijcv}. The class labels from Replica and AVD are both separately mapped to those in ADE20K dataset resulting in 52 and 66 classes respectively. We discard classes that do not have an unambiguous overlap. 



\noindent \textbf{View-Pair Selection } \quad
We heuristically sample informative pairs of images, by considering uniformly sampled views on a grid and selecting neighboring views with varying degree of overlap as characterized by Intersection over Union (IoU) measure. 
The view pairs with IoU in the range of $[iou_{l}, iou_{h}]$ are selected for training. This sampling process reduces computation during training as it needs only be done once per environment for all the experiments.

\subsection{Implementation Details}  \label{sec:exp-implement}

We use a slightly modified version of DeeplabV3+ \cite{Deeplabv3plus} as our segmentation model as shown in Figure \ref{fig:model} with ResNet50 \cite{he_2016_cvpr_resnet} backbone.
We randomly initialize all the weights of the model.

\paragraph{\it Pre-training}
We pre-train the models using all view-pairs sampled as explained in \ref{sec:exp-data} using $iou_l=0.3$ and $iou_r=0.9$. To generate regions, we use the efficient graph based segmentation method \cite{Felzenszwalb_2004_IJCV_EfficientGI} with $scale = 250$ and $sigma = 2000$. We obtained these parameters by calculating IoU with available 5\% ground truth annotation in Replica for various values. We take the output before the final layer as the feature on which the pair-loss is calculated as explained in Section \ref{sec:method-pairloss}. We resize the feature map to original input resolution using nearest neighbor interpolation to project and match across views. We use a batch-size of 16 i.e. 32 pairs of related views by default. In each batch, we sample $|S| = 81920$ pixels. We use strong and weak image augmentations for $I_1$ and $I_2$ respectively, which we found performs better than using strong augmentations for both. 
With Barlow Twins loss we also found that it helpful to use a norm gradient clipping of $5$ and $\lambda = 0.005$ \cite{barlow}. We use a learning rate of 0.01 for 50K iterations with a cosine decay scheduler \cite{Loshchilov_ICLR_2017_sgdr} which decreases the learning rate by a factor of 10. We use a learning rate warm-up period of 2500 iterations. We use a V100 GPU to train the models.

\paragraph{\it Fine-tuning}
For fine-tuning we use ground truth annotation from 5\% (16) of all the images in the Replica environment and remaining image annotations for testing. For AVD, we create 2 sub-datasets. In \textbf{\em AVD-easy}, we choose 5 annotated images for testing that have high overlap with the training set which consists of 38 (1.57\%) of the 2412 images. However, in \textbf{\em AVD-hard}, we fine-tune on 24 (1\%) images randomly chosen while testing on the remaining 19. For fair analysis, we use the same set of images for training and testing across all the experiments. We use a learning rate of 0.01 with a  polynomial scheduler for 80K iterations and a weight decay of $5e^{-4}$.  For the segmentation model trained on ADE20K, we found that using a learning rate of 0.001 is better than 0.01 because the model is already suitable for segmentation out-of-the-box.

\subsection{Baselines}       \label{sec:exp-baselines}
First, we try our approach using exact pixel matching. Specifically, we project points from one image/view of the environment to another as mentioned in section \ref{sec:method-temporal} and train using the full set of corresponding pixel-pairs. While looking at the predictions made by our model, we found that most of the wrongly predicted pixels were assigned to background classes wall or floor. 

\noindent \textbf{Hypotheses}: We hypothesize two possible reasons for the poor performance in exact matching. The first reason is the inherent class imbalance in indoor environments.
The second reason is that the pixel-pairs between the two views have low variability between them i.e. the pixels look similar across the views. In order to verify this hypothesis, we perform experiments by assuming the availability of ground truth labels while pre-training. This gives us an upper bound of our proposed method. We try all combinations of sampling and matching techniques for the pixel-pairs as explained in Section \ref{sec:method-sampling} but using ground truth labels as regions.



The results for the baseline experiments are shown in Table \ref{tab:baselines}. 
Because of the class imbalance, random sampling (RaX, RaR) results in a higher number of samples from large classes which overweight the smaller classes. Balanced sampling (BaX, BaR) is better with either of the matching methods because an equal number of pairs are sampled from each class. This confirms our class-imbalance hypothesis. Region matching (BaR) is better than exact matching (BaX) with balanced sampling . This shows that it is also important to match non-corresponding pixels from the same class. Hence, this validates our low variability hypothesis. We also point out that sampling in a class-balanced way is more important than matching because in balanced sampling method, both matching strategies are higher than those in random sampling. 
Interestingly, when performing random sampling we observe that it is best to match exact projection (RaX). However, matching this way limits the possible number of pairs that can be formed so many possible informative pairs can be missed. This is verified with our best performing model that uses region matching.

\begin{table}
\caption{Supervised Baselines}
\label{tab:baselines}
\begin{center}
\begin{tabular}{l|c|c|c|c}
\hline
Name & Supervision & Sampling & Matching & mIoU \\
\hline
BaR & labels & \textbf{Ba}lanced & \textbf{R}egion & \textbf{73.4} \\
BaX & labels & \textbf{Ba}lanced & e\textbf{X}act & 61.2 \\
RaR & labels & \textbf{Ra}ndom & \textbf{R}egion & 48.6 \\
RaX & labels & \textbf{Ra}ndom & e\textbf{X}act & 60.5 \\
\hline
\end{tabular}
\end{center}
\end{table}

\begin{table}
\caption{Results on Replica}
\label{tab:result}
\begin{center}
\begin{tabular}{l|c|c|c|c}
\hline
Weights         & Model-weights     & Dataset & Pre-Training & mIoU \\
\hline
\hline
Random          & --            & --        & --                &   43.9 \\
\hline
ImageNet        & ResNet50      & ImageNet  & classification        &   48.7 \\
ADE20K          & DeepLabV3+    & ADE20K    & segmentation        &   \underline{59.2} \\
\hline
PixPro1x \cite{PixPro} 
                & ResNet50      & ImageNet  & self-sup   &   53.7 \\
PixPro4x \cite{PixPro} 
                & ResNet50      & ImageNet  & self-sup   &   54.7 \\
\hline
RegConsist           & DeepLabV3+    & Replica        & self-sup                &  \textbf{62.7} \\
\hline
\end{tabular}
\end{center}
\end{table}

\begin{table}
\caption{Results on AVD}
\label{tab:result-avd}
\begin{center}
\begin{tabular}{l|c|c}
\hline
Weights         & mIoU (AVD-easy) & mIoU (AVD-hard) \\ 
\hline
\hline
Random          & 66.7              &       49.1    \\
\hline
ADE20K          & 69.3              &       \textbf{69.1}    \\
\hline
RegConsist (Ours)   & \textbf{69.5}              &       64.8    \\

\hline
\end{tabular}
\end{center}
\end{table}

\subsection{Results} \label{sec:exp-result}

\input{tex_files/result_figures}

We use estimated regions during training and apply our RegConsist approach on Replica dataset. The results are shown in Table \ref{tab:result}. To compare, we change the initialization of DeepLabV3+ with other trained or pre-trained models. ImageNet is the ResNet50 model trained in a supervised manner for image classification on the ImageNet dataset \cite{Deng_CVPR_2009_imagenet}. Similarly ADE20K model was trained on ADE20K dataset \cite{ade20k, ade20kijcv} for semantic segmentation in a supervised manner for 200 epochs which reaches an mIoU of 39.8 in the ADE20K validation set. PixPro1x and PixPro4x \cite{PixPro} were both trained on ImageNet in a self-supervised manner using single view augmentation techniques on the ImageNet dataset for 100 and 400 epochs respectively. Our method performs better than even the DeepLabV3+ model trained for ADE20K dataset which shows that it is important to pre-train the model in a new environment. It still does not reach the ground truth baseline because the estimated regions do not precisely match the ground truth labels.

Similarly, we compare our method with the worst and the best model from Replica, excluding ours, on AVD sub-datasets. The results are shown in Table \ref{tab:result-avd}. AVD-easy is easy even for the Random initialization but AVD-hard is very difficult. RegConsist does not perform better than ADE20K on AVD-hard. We suspect that the gap between AVD and ADE20K is simpler (both are real world with indoor images) so ADE20K is able to utilize its existing knowledge. However the large domain gap makes it difficult to do the same in Replica. This demonstrates that our method is very good in situations where the domain gap is large or when there is no dataset on which supervised learning can be done beforehand. Similarly, our method depends on the precision of the regions we obtain and we did not tune the regions for AVD unlike Replica.


\subsection{Ablations}

\noindent \textbf{Number of Labeled examples} \quad
We experiment by changing the amount of labeled images for fine-tuning. We train for 80K iterations instead of 20K. 
The results are shown in table \ref{tab:num-labeled}. As can be seen, our method performs better than that trained on ADE20K in every case. The gap between the performance decreases the more labeled examples are available. This shows that our model is more suited in the fewer annotation regime.

\begin{table}
\caption{Number of Labeled Examples}
\label{tab:num-labeled}
\begin{center}
\begin{tabular}{l|c|c|c|c}
\hline
Labeled Images      & 5\%       & 10\%      & 20\%      & 30\%   \\
\hline
ADE20K              & 59.2      & 67.2      & 76.8      & 80.1  \\
\hline
RegConsist (Ours)                & \textbf{62.7}      & \textbf{67.8}         & \textbf{77.3}      & \textbf{80.5}  \\
\hline
\end{tabular}
\end{center}
\end{table}

\noindent\textbf{Starting weights} \quad
Instead of starting from scratch during pre-training, we experiment with other weight initialization as shown in Table \ref{tab:start-weights}. Changing the start weights does not affect the performance much because our method is inherently instance specific. Region consistency is limited to matching spatially and visually similar pixels in the images; an object may be covered by one of the regions but two objects from the same class will not have a direct association between them. Hence, any category-specific information obtained from weight initialization is unused and/or lost. 
So the model learns in a bottom-up approach: in the pre-training phase instances are learned followed by categories in the fine-tuning phase.

\begin{table}
\caption{Starting Weights}
\label{tab:start-weights}
\begin{center}
\begin{tabular}{l|c|c|c}
\hline
Initialization & Dataset & Training & mIoU \\
\hline
Random                      & --        & None              & 57.0 \\
ImageNet-sup                & ImageNet  & Supervised        & 56.7 \\
PixPro-400 \cite{PixPro}    & ImageNet  & Self-supervised   &  \textbf{57.4} \\
\hline
\end{tabular}
\end{center}
\end{table}

\begin{figure}
    \centering
    \includegraphics[width=0.5\textwidth]{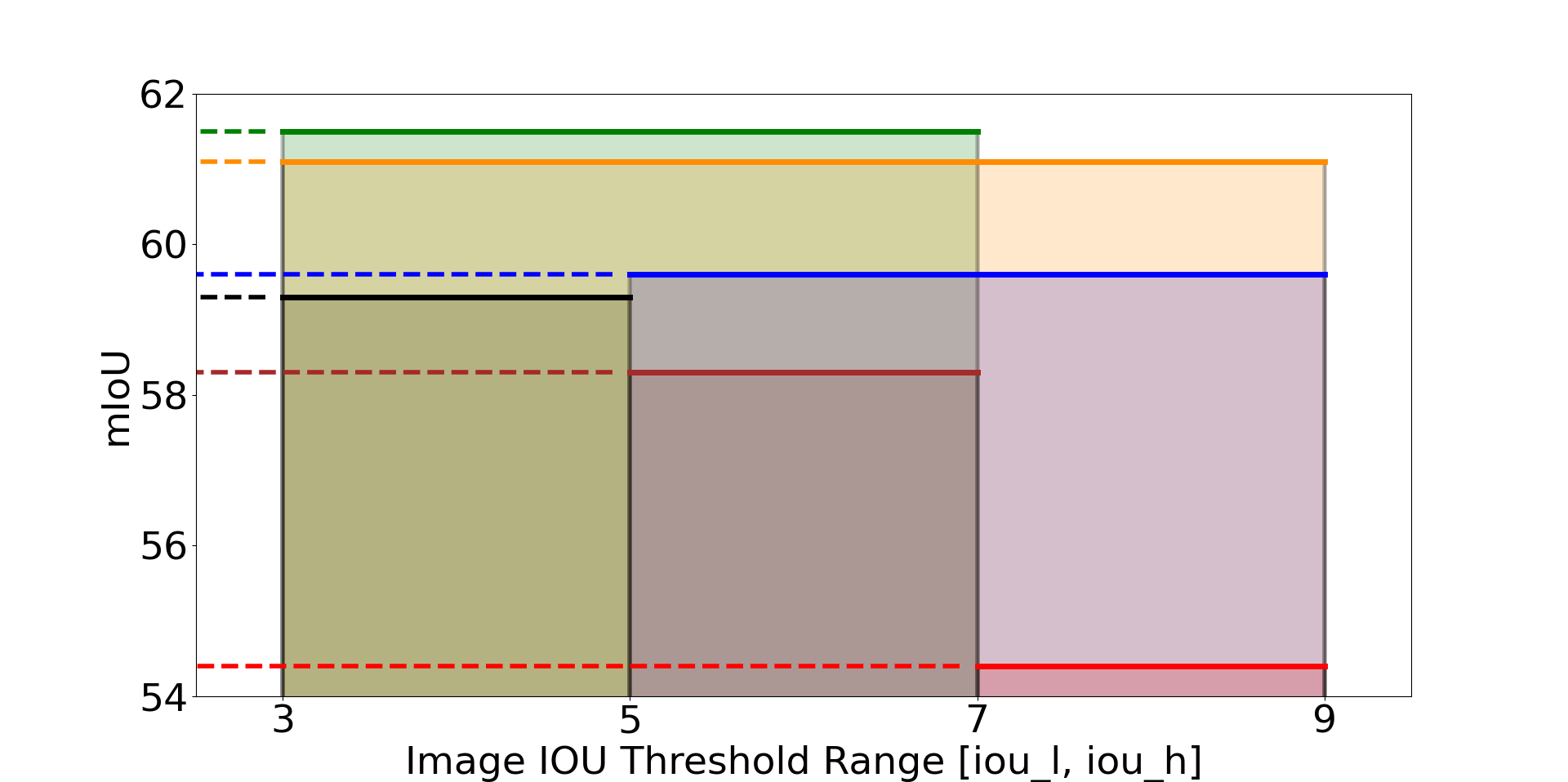}
    \caption{Image IoU threshold range vs accuracy.}
    \label{fig:ablations-image-iou}
\end{figure}

\noindent\textbf{Image IoU threshold} \quad When choosing view-pairs from the environment, we check various IoU threshold ranges by changing the values of $iou_l$ and $iou_h$ and fine-tuning for 20K iterations. From Figure \ref{fig:ablations-image-iou}, we can see that threshold [7-9] produces the worst result of 54.4 because the images in the pairs are very similar to each other, obtained through minor movements of the agent. We find that threshold [3-7] produces the best results which shows that it is important to keep a balance between similar and dissimilar view-pairs.

\section{Conclusion}
We have demonstrated the effectiveness of self-supervised pre-training of semantic segmentation models in an indoors environment by exploiting spatial and temporal consistency between regions in overlapping views. The method exploits the ability to register neighboring views of indoors scenes and uses efficient generation of positive training examples for contrastive learning framework.  
The proposed approach was validated through several experiments and ablation studies, demonstrating the effects of different choices of sampling strategies, amounts of labeled data as well as comparisons with alternative baseline approaches that rely on fine-tuning of models pre-trained on labeled data. 
We also investigate properties of our approach by starting from various weight initialization led us to discover that it gains instance specific knowledge. We argue and show that this level of information helps the model to better learn category specific information in the fine-tuning stage. 
We show that our approach, RegConsist, allows the agent to learn as good as a model trained on labeled  data from a related domain. 
The assumption of availability of labeled images from a similar domain is not valid for all possible domains. 
Our approach is especially beneficial in such scenarios. 










\bibliographystyle{IEEEtran}
\bibliography{IEEEabrv,mybibfile}

\end{document}


\maketitle
\thispagestyle{empty}
\pagestyle{empty}

\subsection{Image Augmentations}
We use strong and weak image augmentations as shown in Table \ref{tab:supp-aug}. For AVD dataset \cite{Ammirato_ICRA_2017_AVD}, we change the RandomResizedCrop scale from (0.08, 1.0) to (0.5, 1.0) because
AVD contains large areas with depth holes along the left and right edges of the image so there might not be enough points to project with stronger cropping along those areas.

\begin{table}
\caption{Pixel-Pairs Batch Size}
\label{tab:supp-aug}
\begin{center}
\begin{tabular}{|l|c|c|c|}
\hline
Augmentation        & Parameter & Strong & Weak \\
\hline \hline
RandomResizedCrop   & size      &   224         &   224 \\
RandomResizedCrop   & scale     &   (0.08, 1.0)  &   (0.9, 1.0)  \\
RandomResizedCrop   & ratio     &   (0.75, 1.25)  &   (0.75, 1.25)  \\
\hline
RandomColorJitter   & brightness     &   0.3  &   --  \\
RandomColorJitter   & contrast     &   0.3  &   --  \\
RandomColorJitter   & saturation     &   0.3  &   --  \\
RandomColorJitter   & hue     &   0.15  &  --  \\
RandomColorJitter   & p     &   0.8  &  --  \\
\hline
RandomGrayScale		&	p		&  0.2		& -- \\
\hline
RandomGaussianBlur	&	p		&  0.5		& -- \\
\hline
\end{tabular}
\end{center}
\end{table}

\subsection{More Ablations}
\noindent\textbf{Pixel-Pairs Batch Size}
When using regions, it is possible to sample a variety of number of positive pixel-pairs $|S|$ up to the all possible combinations of pixels within each matched regions across views. We try various values for $|S|$ and show the results in table \ref{tab:batch-pixel}. \textit{Possibly plot graph instead of table.}

\begin{table}
\caption{Pixel-Pairs Batch Size}
\label{tab:batch-pixel}
\begin{center}
\begin{tabular}{c|c|c|c|c|c}
Size & 1x & 2x & 4x & 20x & 80x \\
\hline
mIoU & 1x & 2x & 4x & 20x & 80x \\
\hline
\end{tabular}
\end{center}
\end{table}

\subsection{View Sampling in Replica dataset} \label{supp:view-sampling}
In order to ease the training process, we pre-sample views from the Replica environment. We load Replica \cite{replica19arxiv} in Habitat simulator \cite{habitat} which allows us to simulate an agent and sample views along with their semantic segmentation labels and depth values. We randomly sample a fixed height that remains constant throughout the sampling process. We uniformly sample views from multiple locations in a grid-like manner where each grid-cell is 1 unit apart in both the axes of ground-parallel horizontal plane. From each such grid-cell, we sample views 45 degrees apart from each other, around the vertical axis with 1024 x 1024 resolution. Sampling this way results in around 312 images or views from the environment.

\subsection{Classes mapped from Replica to ADE20K}

Classes are mapped from Replica to ADE20K in a many to one fashion i.e. multiple classes from Replica can be mapped to a single class in ADE20K. If any class is ambiguous in that it can be mapped to more than one class in ADE20K, it is ignored. Any class that does not have a mapping to ADE20K are also ignored. The remaining classes are shown in Table \ref{tab:supp-replica-ade-classes}.

\begin{table*}
\caption{Supervised Baselines}
\label{tab:supp-replica-ade-classes}
\begin{center}
\begin{tabular}{|l|l|l|l||l|l|l|l|}
\hline
Label & Replica & Label & ADE20K & Label & Replica & Label & ADE20K\\
 \hline
    2 &  base-cabinet    &   11 & cabinet         & 44 &  indoor-plant    &   18 & plant;flora;pla \\ 
  3 &  basket          &  113 & basket;handbask & 47 &  lamp            &   37 & lamp            \\
  4 &  bathtub         &   38 & bathtub;bathing & 50 &  mat             &   29 & rug;carpet;carp \\ 
  7 &  bed             &    8 & bed             & 51 &  microwave       &  125 & microwave;micro \\ 
  8 &  bench           &   70 & bench           & 59 &  picture         &   23 & painting;pictur \\ 
  9 &  bike            &  128 & bicycle;bike;wh & 60 &  pillar          &   43 & column;pillar   \\ 
 10 &  bin             &  139 & ashcan;trash;ca & 61 &  pillow          &   58 & pillow          \\ 
 11 &  blanket         &  132 & blanket;cover   & 63 &  plant-stand     &  126 & pot;flowerpot   \\ 
 12 &  blinds          &   64 & blind;screen    & 64 &  plate           &  143 & plate           \\ 
 13 &  book            &   68 & book            & 67 &  refrigerator    &   51 & refrigerator;ic \\ 
 14 &  bottle          &   99 & bottle          & 69 &  scarf           &   93 & apparel;wearing \\ 
 15 &  box             &   42 & box             & 70 &  sculpture       &  133 & sculpture       \\ 
 18 &  cabinet         &   11 & cabinet         & 71 &  shelf           &   25 & shelf           \\ 
 20 &  chair           &   20 & chair           & 73 &  shower-stall    &  146 & shower          \\ 
 22 &  clock           &  149 & clock           & 74 &  sink            &   48 & sink            \\ 
 24 &  clothing        &   93 & apparel;wearing & 76 &  sofa            &   24 & sofa;couch;loun \\ 
 29 &  cushion         &   40 & cushion         & 78 &  stool           &  111 & stool           \\ 
 30 &  curtain         &   19 & curtain;drape;d & 80 &  table           &   16 & table           \\ 
 31 &  ceiling         &    6 & ceiling         & 84 &  toilet          &   66 & toilet;can;comm \\ 
 33 &  countertop      &   71 & countertop      & 86 &  towel           &   82 & towel           \\ 
 34 &  desk            &   34 & desk            & 87 &  tv-screen       &   90 & television;tele \\ 
 36 &  desktop-compute &   75 & computer;comput & 91 &  vase            &  136 & vase            \\ 
 37 &  door            &   15 & door;double;doo & 93 &  wall            &    1 & wall            \\ 
 40 &  floor           &    4 & floor;flooring  & 96 &  wardrobe        &   36 & wardrobe;closet \\ 
 43 &  handrail        &   96 & bannister;banis & 97 &  window          &    9 & windowpane;wind \\ 
 98 &  rug           &   29 & rug;carpet;carp & 100 &  bag            &  116 & bag             \\

\hline
\end{tabular}
\end{center}
\end{table*}

\subsection{Region Matching Algorithm}

When matching regions across image-pairs $I_1$ and $I_2$, we need to calculate the iou of each region projected from $I_1$ to $I_2$ with the regions independently calculated in $I_2$. It is not efficient to do this iteratively for each possible pairs in brute force method. Therefore, we devise a new method to calculate the iou that uses a pairing function such as Hopcroft and Ullman pairing function or Cantor pairing function. A pairing function is a bijective function that maps unique pairs of integers to a single integer with one-to-one correspondence. We use Cantor pairing function because it works with any non-negative integers. 

\begin{algorithm}
\caption{Efficient Region Matching}\label{alg:reg-matching}
\begin{algorithmic}[1]
\Require $I_1 = \{u_i^k\}, I_2 = \{v_i^k\}$
\Ensure No two regions have same label in a single image
\State Find unique regions and their count in U and V
\State $O \gets $ Pair images U and V element-wise
\State $\hat{O} \gets $ Unique regions in $O$ and their count

\For{$i$ in $\hat{O}$}
    \State (
\If{$N$ is even}
    \State $X \gets X \times X$
    \State $N \gets \frac{N}{2}$  \Comment{This is a comment}
\ElsIf{$N$ is odd}
    \State $y \gets y \times X$
    \State $N \gets N - 1$
\EndIf
\EndFor
\end{algorithmic}
\end{algorithm}

\textbf{More Examples of Semantic Segmentation} \\
\textbf{Qualitative examples on AVD} \\
\textbf{Replica Classes and ADE20K mapping}









\bibliographystyle{IEEEtran}
\bibliography{IEEEabrv,mybibfile}

%% file: tex_files/Introduction.tex
\section{Introduction}


Semantic segmentation has been used extensively for both semantic mapping ~\cite{Cesar_SLAM16} and also as input representation for training policies for embodied agents (e.g. policies for target driven or point goal navigation) that rely on visual perception~\cite{Chaplot_NEURIPS_2020_object_goal_nav, georgakis_ICLR_2022_active_learning_semantic_nav}. 
Training semantic segmentation model for a particular environment requires a large amount of per-pixel annotations~\cite{ade20k} that is very costly and laborious. Alternatively for similar classes of environments that share a large subset of semantic labels, a model can be trained for the entire domain (say indoors environments) followed by domain adaptation~\cite{pmlr-v80-hoffman18a}. 
In a robotic setting the agent is often able to move around and capture large amounts of visual data and the ability to estimate ego-motion and depth perception enables the agent to effectively associate multiple views of the same scene.


\begin{figure}
    \centering
    \includegraphics[width=0.4\textwidth]{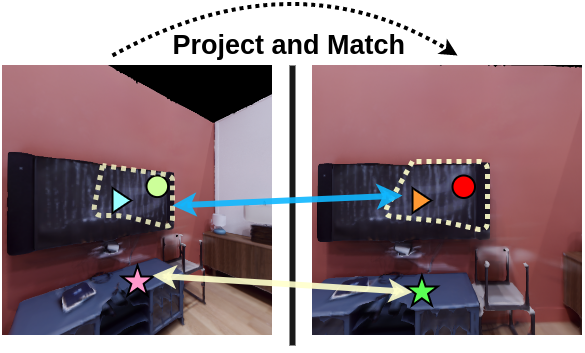}
    \caption{Example of consistency. A pair of views from different location and pose of the agent inside an environment in Replica Dataset \cite{replica19arxiv}. We find and match pixels across the view-pairs. In exact matching correspondence points are matched (yellow arrow). In region matching, any pixel across overlapping regions from the two views can be matched (blue arrow). Best viewed digitally or in color.}
    \label{fig:model}
\end{figure}

Recent years marked notable progress in various self-supervised methods for training large DCNN's from scratch without relying on commonly used backbones pre-trained on ImageNet. The existing techniques used various forms of contrastive learning and different pretext tasks such as predicting the masked portion of the image \cite{He_CVPR_2022_Masked_encoders}, predicting masks of objects  \cite{henaff_arxiv_2022_object_discovery} or predicting the rotation of the image \cite{gidaris_2018_iclr_pretrain_rotation}. The self-supervised pre-training is then followed by fine-tuning with a small fraction of the labeled data.  Even though these models have been proven to be very effective, they are typically evaluated for image classification tasks.

In this paper we explore the use of self-supervision that comes from the spatial and temporal consistency between pairs of overlapping views and demonstrate how to pre-train Deep Convolutional Neural Network (DCNN) model for semantic segmentation using data captured in the environment of interest.
We assume that within a single traversal path, the environment remains static so as to simplify the process of computing correspondences between neighboring views, that will be used for self-supervised training of the model. 

\noindent {\bf Contribution}
1) We propose RegConsist, a method for self-supervised pre-training of a semantic segmentation model using spatial and temporal consistency cues. We exploit correspondences between multiple views for generating positive examples for contrastive learning framework and evaluate the effect of different sampling strategies on the result.
2) We demonstrate that the resulting model can be fine-tuned with only a small fraction of image annotations, obtaining competitive or better performance in a novel indoor environment compared to fully supervised models. 3) We are the first to demonstrate that the Barlow twins loss~\cite{barlow} works well for semantic segmentation in an indoor environment. 4) We demonstrate the efficacy of our method on Replica ~\cite{replica19arxiv} and AVD \cite{Ammirato_ICRA_2017_AVD} datasets both qualitatively and quantitatively while using as low as 5 percent of the annotated data. 5) We perform extensive ablation studies to verify our method's performance including different starting conditions of the segmentation model.



\begin{figure*}
\begin{center}

\includegraphics[width=0.8\textwidth]{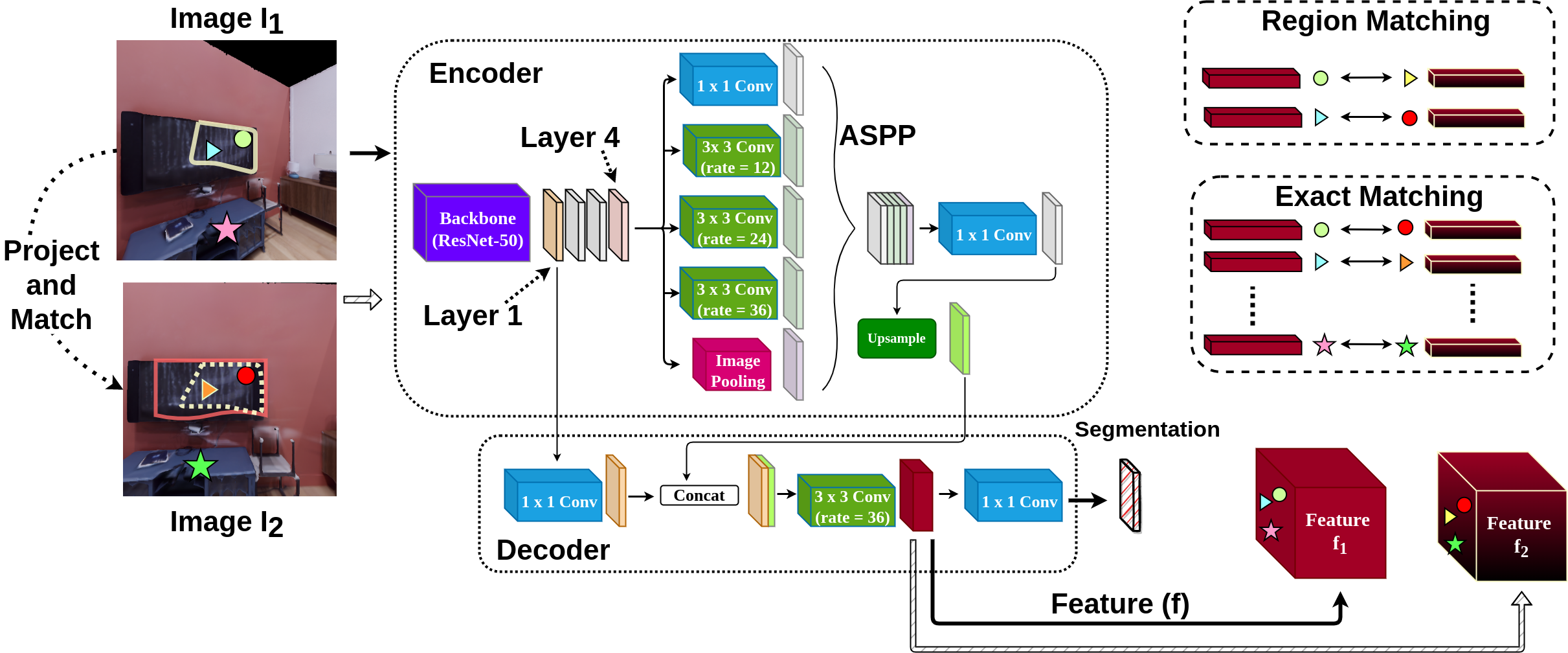}
\end{center}
   \caption{Our proposed method. The segmentation model (DeepLabV3+ \cite{Deeplabv3plus}) separately processes two views that capture the same part of the environment. Correspondence points are obtained from the two views. In exact matching only the correspondence points form positive pixels pairs. In region matching, regions are estimated for each view separately (yellow in $I_1$ and red in $I_2$). Then, points are matched from highly overlapping estimated regions across the views (in $I_2$, red region with dotted yellow region projected from $I_1$). The positive pixel-pairs are aligned using Barlow Twins loss \cite{barlow}. Best viewed digitally or in color.}
    \label{fig:model}
\label{fig:short}
\end{figure*}

%% file: tex_files/RelatedWork.tex
\section{Related Work}

Our work relates to previous self-supervised learning  efforts 
in robotics, exploiting the idea that robots are able to gather their own data 
and with proper training strategies do not have to rely on laborious labeling 
required for supervised training. In the past this
idea was explored in the context of different tasks, such as pose-estimation~\cite{FoxPose_ICRA20}, 
object detection~\cite{selfsupervisedPot_2018,Frakiadaki_BMVC21} 
and learning object specific visual representation 
for manipulation and tracking~\cite{Tedrake_CORL20},
to mention a few. Reinforcement learning and contrastive learning were the two most commonly 
used frameworks that spear-headed the technical advances in self-supervised learning 
in both computer vision and robotics. Below we review in more detail the works that are 
most relevant to our task of semantic segmentation. 

\subsection{Self-supervised Learning}
To mitigate the need for large amounts of labeled data required for training deep models, self-supervised methods typically use various pretext tasks to generate training data. In the past, in single image setting, these 
included masked image modeling \cite{He_CVPR_2022_Masked_encoders}, object mask prediction \cite{henaff_arxiv_2022_object_discovery}, instance discrimination \cite{zhirong_wu_2018_CVPR_instance_discrimination_memorybank} and others. These auxiliary tasks provide the model with the desired objective to embed semantically similar inputs closer in the learned embedding space. 

\paragraph{\it Contrastive Learning}
Contrastive learning \cite{contrastive} has been a "workhorse" of self-supervised learning approaches for training DCNNs. It exploits the ability to associate (semantically) similar examples (positive pairs) and distinguish them from negative pairs as a supervision signal for learning suitable representations (embeddings). The existing approaches vary depending on the final task, CNN architectures (often using variations of Siamese Neural Network architectures), loss functions and methods for obtaining similar and dissimilar training examples. The most common concern in representation learning is to avoid model collapse. Authors in~\cite{zhirong_wu_2018_CVPR_instance_discrimination_memorybank} consider each image as a separate class and the model is trained to disambiguate an image from other images in the dataset. MoCo~\cite{MoCo} employed a dynamic memory bank \cite{zhirong_wu_2018_CVPR_instance_discrimination_memorybank} to store features from current the iteration of the model as negatives during training.  In order to remove the requirement of memory bank, SimCLR \cite{SimCLR} proposed to use negative pairs from the mini-batch itself but consequently necessitating a larger batch size. BYOL \cite{BYOL} further remove the need for the negative examples altogether, by introducing asymmetric architecture. Recently, \cite{barlow} introduced a new objective function termed Barlow Twins, based on redundancy reduction, which removes the need for large batches and asymmetric models altogether. 

\paragraph{\it Point Level Contrast.} While contrastive learning at the global image level has proven to be beneficial for image classification, the problems such as object detection and semantic segmentation where predictions are done at the object bounding box or pixel level, require disambiguation of finer features. 
Getting pixel level positive pairs is more difficult than their image level counterpart. PixPro~\cite{PixPro} follow SimSiam~\cite{SimSiam} like training but at the pixel level; the positives are obtained by thresholding the cosine distance between the features at the pixel level within a given image. 
Authors in~\cite{LookBeyond} sample positive pixel pairs within regions obtained by k-means clustering of the initial features, 
while~\cite{Bai_CVPR_2022_point_level_region_contrast} circumvent the need of finding regions by dividing the image into a fixed N x N grid where each grid-cell is considered a separate region. 

In situations where a small number of labeled examples is available one can approach learning using both cross entropy loss based on available labels along with self-supervised loss. This requires a momentum-updated teacher network along with a memory bank to associate features between examples outside of the images in a given iteration \cite{Alonso_2021_ICCV}. 
In order to obtain additional labels if labels are sparse~\cite{SemanticFusion,SemiSegMesh} proposed to use label propagation techniques. This however requires complete and accurate 3D reconstruction in order to fuse predictions from different frames.

Our work is most closely related to the efforts of self-supervised learning for object detection~\cite{Frakiadaki_BMVC21,selfsupervisedPot_2018} by using multiple views and
view association to guide the training, but extends these ideas to dense pixel-level prediction 
tasks such as semantic segmentation. 

%% file: tex_files/Method.tex
\section{Method}  \label{sec:method}
We assume a robotic agent with the ability to perceive and recover ego-motion and 3D structure of the
environment and associate overlapping views of the same scene. This can be achieved with appropriate sensors such as a depth sensor or stereo camera or with the use of 3D structure and motion estimation techniques \cite{kosecka_2003_3dvision} or suitable SLAM approach~\cite{Cesar_SLAM16}.
To instantiate a self-supervised learning approach for semantic segmentation we propose a (\textbf{Reg}ion \textbf{Consist}ency)
\textbf{RegConsist} method for temporal and spatial alignment of overlapping views that we describe next.


 
\subsection{Temporal Consistency} \label{sec:method-temporal}
Let $I_{1}$ and $I_{2}$ be two images captured by the agent in the fixed indoor environment. 
Assuming the availability of known intrinsic and extrinsic camera parameters and depth, we can associate the pixels in the overlapping views of the same scene using (\ref{eq:project}).
\begin{equation}
    T_{1 \rightarrow 2}(I_1) = \{ K ( {T_2}^{-1} (T_1( K^{-1}(\mathbf{x})) \quad \forall  \mathbf{x} \in I_1 \} \label{eq:project}
\end{equation}
where, $K$ is the intrinsic parameters of the camera, $T_1 = [R_1|t_1]$ is the camera pose for the image $I_1$ having rotation $R_1$ and translation $t_1$ with respect to a fixed coordinate system and $\mathbf{x}$ is a pixel in $I_1$. The operator $T_{1 \rightarrow 2}$ transforms the 2D pixel coordinates in $I_1$ to the 3D world coordinate system and projects it back to pixel coordinates in $I_2$. We assume that reliable correspondences can be estimated either using learning based method~\cite{Truong_2021_ICCV_unsupervised_correspondence} or, as in our case, with availability of the depth sensor. 
Let $I_1^p$ and $I_2^q$ be $p^{th}$ and $q^{th}$ pixels in the images $I_1$ and $I_2$ respectively. If the pixels belong to the same 3D location in the environment and unoccluded in both images, they are semantically identical. Let $S_t=\{(p,q)\ |\ p \in I_1, q \in I_2 \}$ be the set of all such pixel-pairs. We consider all pairs in $S_t$ as positives and enforce their features to be aligned across the views. 

\subsection{Spatial Consistency} \label{sec:method-spatial}
The pixel-pairs we obtain from Section \ref{sec:method-temporal} may not have high variability, {\em  i.e.} they might look very much similar across views, especially when views are close. Given the pixel $p$ in $I_1$ and a small neighborhood region $\mathcal{N}(p)$ around the pixel, all pixels $\hat{p}$ inside the region are highly likely to belong in the same class as $p$ (yellow region in Figure \ref{fig:model}). Similarly in $I_2$, all $\hat{q} \in \mathcal{N}(q)$ belong to same class as $q$ (dotted yellow region in $I_2$) . Hence any pixel in $N(p)$ can be matched with any pixel in $N(q)$. Therefore, we also consider such pixel-pairs to be positives. Specifically, $S = \{(\hat(p),\hat{q}) | p \in I_1, \hat{p} \in \mathcal{N}(p), q \in I_2, \hat{q} \in \mathcal{N}(q)\}$. It is important to note that $S_t \subset S $ because $p \in \mathcal{N}(p)$ and $q \in \mathcal{N}(q)$. Hence, the pairs are enforced to be aligned across overlapping and consistent regions (\textbf{Reg}ion \textbf{Consist}ency).

Regions can be estimated using unsupervised methods such as the efficient graph based segmentation method \cite{Felzenszwalb_2004_IJCV_EfficientGI}. Because the regions are independently estimated across the views, they are not perfectly aligned. So to get highly overlapping regions, we need to find IoU between each region in $I_1$ with those in $I_2$. This requires $O(|R_1| . |R_2|)$ where $|R_1|$ and $|R_2|$ are the number regions in $I_1$ and $I_2$. This is extremely slow to calculate in each iteration and hampers training speed. Therefore we devise a new algorithm to calculate this in $O(|R_1 \cap R_2|)$ where $|R_1 \cap R_2|$ is the number of overlapping regions between $R_1$ and $R_2$ that can be computed using the Cantor pairing function.

\subsection{Pixel-Pair Sampling and Matching} \label{sec:method-sampling}


While using all the pairs from the set $S$ is possible, it is not an efficient method. So, we sample pixel-pairs from $S$ in each batch. We proceed by first sampling the first pixel $p$ from image $I_1$. We then follow up by matching a suitable pixel $q$ from image $I_2$ which is either spatially or temporally consistent as explained in Sections \ref{sec:method-temporal} and \ref{sec:method-spatial}.

In \textbf{\em random} sampling, we sample pixel $p$ uniformly across the whole image $I_1$. However, in indoor environments, class imbalance exists between large classes such as wall or floor and small classes such as cushion or plate as shown in Figure \ref{fig:segmentation}. To overcome this problem, in \textbf{\em balanced} sampling, we sample pixel $p$ uniformly per region such that there are equal number of points from each region within a batch.

Once the first pixel $p$ in the pair has been sampled from a view $I_1$, we need to match it with a positive pixel from $I_2$. In \textbf{\em exact} matching, we match the $p$ with pixel $q$ which is the exact correspondence of $p$ that satisfies (\ref{eq:project}). To get variability between the pixels in the positive-pair, in \textbf{\em region} matching, we match $p$ with $q$ sampled uniformly from region $R_2$ that matches region $R_1$ in $I_1$ in which $p$ resides, as explained in Section \ref{sec:method-spatial}.


\subsection{Pair Loss}  \label{sec:method-pairloss}

Let $|S|$ be the total number of pairs sampled in a batch. Let $f_1^p$ and $f_2^q$ be the features of pixels $p$ and $q$ obtained from images $I_1$ and $I_2$ respectively. For brevity, we overload the notation $p$ and $q$ to represent pixels as well as their features $f_1^p$ and $f_2^q$ respectively. In order to align the features of pixel-pairs, we use Barlow Twins loss \cite{barlow} given by equation (\ref{eq:loss-barlow}). 
\begin{equation}
    \mathcal{L}_{pair} = \sum_i (1-\mathcal{C}_{ii})^2 + \lambda \sum_i \sum_{j \neq i} {\mathcal{C}_{ij}}^2 \label{eq:loss-barlow}
\end{equation}

\noindent where $\mathcal{C}$ is the cross-correlation matrix computed between features $p$ and $q$ in the set $S$ computed along the batch dimension and is given by equation (\ref{eq:barlow-cross}).

\begin{equation}
    \mathcal{C}_{ij} = \frac{ \sum_b p_{i}^{(b)} q_{j}^{(b)} } 
                        { \sqrt{ \sum_b {(p_{i}^{(b)})}^2  }  \sqrt{  \sum_b {(q_{j}^{(b)})}^2  } }
    \label{eq:barlow-cross}
\end{equation}
\noindent where $b$ indexes the batch of pixel-pairs and $i$ and $j$ index the vector dimensions of the features. The first term in the loss in (\ref{eq:loss-barlow}) aligns the input feature-pairs while the second term minimizes the redundancy between each dimension of the features. More details can be found in \cite{barlow}. 

While other contrastive losses can also be used, they require negative pairs or large batch-sizes or momentum encoders which we want to avoid for the sake of simplicity and efficiency. We also tried with SimSiam \cite{SimSiam} like architectures and losses but initial experiments did not work.


\subsection{Supervised Loss}

When using labeled images for fine-tuning we use focal loss \cite{focal} to train the model only on the set of images that have labels. Using focal loss helps mitigate the problem of naturally present class imbalance in indoor environments. 



%% file: tex_files/result_figures.tex
\begin{figure*}
    \begin{center}
        \begin{subfigure}[b]{0.15\textwidth}
            \centering
            \includegraphics[width=\textwidth]{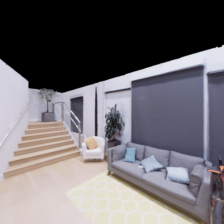}
        \end{subfigure}
        \hfill
        \begin{subfigure}[b]{0.15\textwidth}
            \centering
            \includegraphics[width=\textwidth]{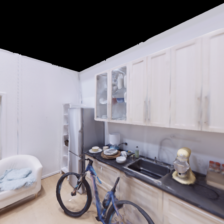}
        \end{subfigure}
        \hfill
        \begin{subfigure}[b]{0.15\textwidth}
            \centering
            \includegraphics[width=\textwidth]{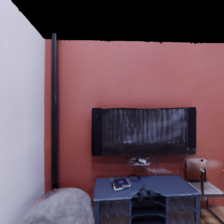}
        \end{subfigure}
        \hfill
        \begin{subfigure}[b]{0.2\textwidth}
            \centering
            \includegraphics[width=\textwidth]{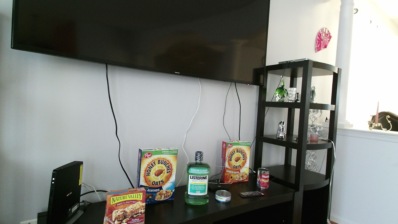}
        \end{subfigure}
        \hfill
        \begin{subfigure}[b]{0.2\textwidth}
            \centering
            \includegraphics[width=\textwidth]{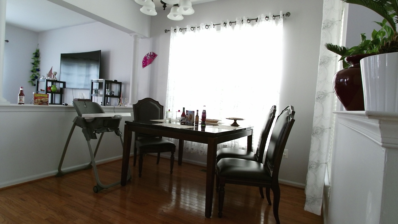}
        \end{subfigure}
        \par\medskip
        \begin{subfigure}[b]{0.15\textwidth}
            \centering
            \includegraphics[width=\textwidth]{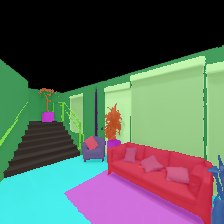}
        \end{subfigure}
        \hfill
        \begin{subfigure}[b]{0.15\textwidth}
            \centering
            \includegraphics[width=\textwidth]{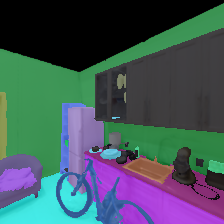}
        \end{subfigure}
        \hfill
        \begin{subfigure}[b]{0.15\textwidth}
            \centering
            \includegraphics[width=\textwidth]{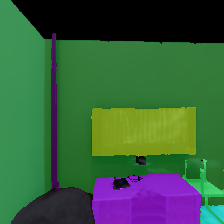}
        \end{subfigure}
        \hfill
        \begin{subfigure}[b]{0.2\textwidth}
            \centering
            \includegraphics[width=\textwidth]{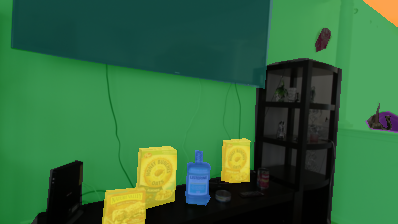}
        \end{subfigure}
        \hfill
        \begin{subfigure}[b]{0.2\textwidth}
            \centering
            \includegraphics[width=\textwidth]{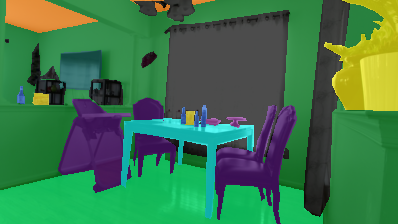}
        \end{subfigure}
        \par\medskip
        \begin{subfigure}[b]{0.15\textwidth}
            \centering
            \includegraphics[width=\textwidth]{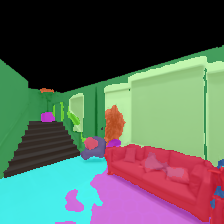}
        \end{subfigure}
        \hfill
        \begin{subfigure}[b]{0.15\textwidth}
            \centering
            \includegraphics[width=\textwidth]{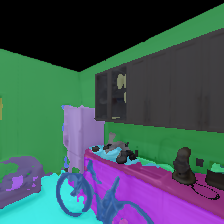}
        \end{subfigure}
        \hfill
        \begin{subfigure}[b]{0.15\textwidth}
            \centering
            \includegraphics[width=\textwidth]{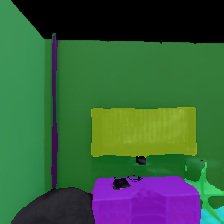}
        \end{subfigure}
        \hfill
        \begin{subfigure}[b]{0.2\textwidth}
            \centering
            \includegraphics[width=\textwidth]{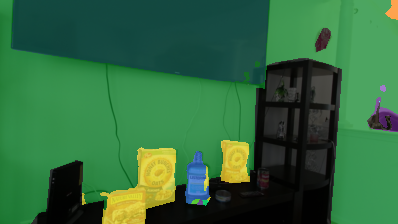}
        \end{subfigure}
        \hfill
        \begin{subfigure}[b]{0.2\textwidth}
            \centering
            \includegraphics[width=\textwidth]{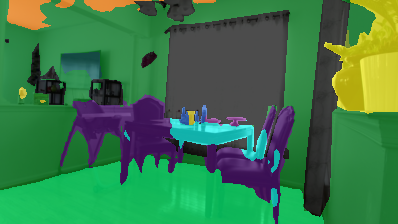}
        \end{subfigure}
        \par\medskip
    \end{center}

\caption{RGB images (top row), ground truth (middle row) and predictions from RegiCon (bottom row). Images from the Replica dataset \cite{replica19arxiv} (3 leftmost columns) and AVD dataset \cite{Ammirato_ICRA_2017_AVD} (2 rightmost columns). Dark pixels in segmentation images do not have a valid class.}
\label{fig:segmentation}
\end{figure*}